\documentclass[conference]{IEEEtran}

\usepackage[pdftex]{graphicx}
\usepackage{amsmath, subfig, caption}
\usepackage{eqparbox}
\usepackage{multirow,tabulary,booktabs}
\usepackage{makecell, booktabs, multirow, siunitx, xcolor}

\begin{document}

%+++++++++++++++++++++++++++++++++++++++++++++++++++
\title{\LARGE Video Contents Understanding using Deep Neural Networks}

%+++++++++++++++++++++++++++++++++++++++++++++++++++

\author{\IEEEauthorblockN{Mohammadhossein Toutiaee\IEEEauthorrefmark{1},
Abbas Keshavarzi\IEEEauthorrefmark{2}, Abolfazl Farahani\IEEEauthorrefmark{3} and
John A. Miller\IEEEauthorrefmark{4}}
\IEEEauthorblockA{Department of Computer Science,
University of Georgia\\
Athens, GA, USA\\
Email: \IEEEauthorrefmark{1}hossein@uga.edu,
\IEEEauthorrefmark{2}abbas@uga.edu,
\IEEEauthorrefmark{3}a.farahani@uga.edu,
\IEEEauthorrefmark{4}jam@cs.uga.edu}}

\maketitle
\thispagestyle{plain}
\pagestyle{plain}

% ================
% # Abstract     #
% ================

\begin{abstract}
We propose a novel application of
\textit{Transfer Learning} to classify video-frame sequences over multiple classes. This is a pre-weighted model that does not require to train a fresh CNN. This representation is achieved with the advent of “deep neural network” (DNN), which is being studied these days by many researchers. We utilize the classical approaches for video classification task using object detection techniques for comparison, such as ``Google Video Intelligence API'' and this study will run experiments as to how those architectures would perform in foggy or rainy weather conditions. Experimental evaluation on video collections shows that the new proposed classifier achieves superior performance over existing solutions. 
\end{abstract}


% ========================
% # I. Introduction      #
% ========================

\section{Introduction}

Traffic Management System (TMS)\cite{chan2005classification} is a field in which the technology is integrated to improve the flow of vehicle traffic and safety. Real-time traffic data from cameras, speed sensors and loop detectors are just few means of monitoring, out of many, to manage traffic flows. Among those, CCTV camera or video surveillance technology is being used more, since it is less expensive and more manageable compared to other ones. Moreover, a fully automatic monitoring system is the interest of certain researchers. 
Majority of the existing framework in monitoring traffic uses sophisticated equations with a substantial number of parameters and coefficients. Antoni B. Chan et al. in \cite{chan2005classification} and \cite{chan2005probabilistic} described methods that are heavily based on motion analysis and object segmentation using ``auto-regressive stochastic'' technique and ``KL-SVM'' classifier, respectively. Both articles showed techniques that are estimated, tuned and utilized once the objects are segmented. One drawback of such techniques is they detect objects in videos when the quality of the frames are visually accurate and the frames are not degraded due to the severe weather conditions, nor distorted due to the corrupted signal. Another issue is that videos should be fed into the proposed models for video classification task, and this requires new weights and parameters estimation specified during the training and tuning process. Last, as Fig~\ref{classes} depicts, Medium and Heavy classes are very similar to one another in terms of appearance and number of vehicles in the videos. These difficulties affect the classifier negatively, and traffic mode prediction based on Entity Detection approach in videos would fail, and  distinguishing between Medium and Heavy classes becomes more challenging. The latter experiment will also be discussed and elaborated in this study along with cases how they would stall. At the end, this article will address those aforementioned problems through suggesting a novel approach, so that video classification case is able to be processed while running over the ``foggy'' weather or ``corrupted'' video frames, and the classification in all traffic modes would be achievable with high accuracy. Additionally, training a new model from scratch over the samples would not be required in this setting. These are the contributions studied via a series of experiments that are elaborated and addressed in the following sections. 

% ==========================
% # III. traffic sys
% ==========================
% \begin{figure}[!ht]
% \centering
% \includegraphics[width=\linewidth]{atms.jpg}
% \caption*{\small{source: www.parknsecure.com}}
% \label{fig:traffic}
% \end{figure}

% ==========================
% # III. Previous work
% ==========================

\section{Previous Work}
\subsection*{Statistical Methods:}
Statistical methods have been practicing by many researches in the context of classification problems. \cite{chan2005probabilistic} presented the ``Dynamic Texture Model'' based on KL classification framework which extracts motion information from the video sequence to determine the motion classes. \cite{chan2005classification} proposed an ``Auto-Regressive Stochastic Processes'' and it was claimed that it would not require segmentation or tracking in the videos to capture the traffic flow. \cite{vaghasia2018approach} presented a hybrid approach that combines the ARIMA model with fuzzy wavelet transform to manage noise attached to dataset that was previously studied. \cite{peng2019multi} compared different statistical and machine learning models including seasonal ARIMA, seasonal VARMA, exponential smoothing and regression, Support Vector Regression, feed forward Neural Networks, and Long Short-Term Memory Neural Networks. Both later articles attempted to forecast of traffic flow in both the short and long terms. One disadvantage of all aforementioned techniques previously studied is they are heavily relying on estimating parameters in a fully supervised fashion, which requires extensive parameters setting and models fitting.

% ========================
% # IV. Entity Detection #
% ========================

\subsection*{CNN Methods:}
Many deep learning frameworks and architectures are being utilized by researchers for different applications and domains and have achieved remarkable results in various computer vision tasks. \cite{szegedy2015going} developed a deep convolutional neural network architecture for image classification task by utilization of the computing resources inside the network. \cite{simonyan2014very} showed a very deep convolutional networks for large scale image classification. \cite{bertasius2015deepedge} presented a multi-scale deep network for image segmentation task. \cite{he2016deep} empirically claimed that their residual networks are easier to optimize, while keeping the accuracy relatively high. \cite{xie2017aggregated} described a highly modularized network architecture with fewer hyper-parameters to set by repeating a building block that aggregates a set of transformations.
\cite{toshev2014deeppose} applied Deep Neural Networks for human pose estimation.
\cite{bertasius2017convolutional} introduced Random Walk Networks (RWNs) for the purposes of object localization boosting and the segmentations that are spatially disjoint. With the advent of CNN era, many scientists research further to deploy DNNs into video-content dataset.
\cite{Ng_2015_CVPR} and \cite{simonyan2014two} implemented a video classification task using stacked video frames as input to the network. \cite{Karpathy_2014_CVPR} studied the performance of CNNs in large-scale video classification and they achieved the highest transfer learning performance by retraining the top 3 layers of the network. \cite{srivastava2015unsupervised} compared and analyzed different proposed models based on LSTMs.

% ====================
% # Video Preparation
% ====================

\section{Video Format and Preparation}

\begin{figure}[ht!]
\centering
\subfloat[Low\label{L}]{%
       \includegraphics[width=0.33\linewidth]{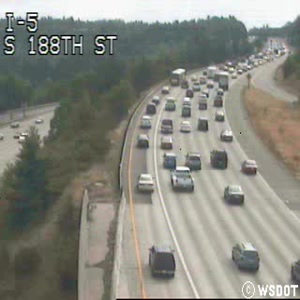}}
\subfloat[Medium\label{M}]{%
        \includegraphics[width=0.33\linewidth]{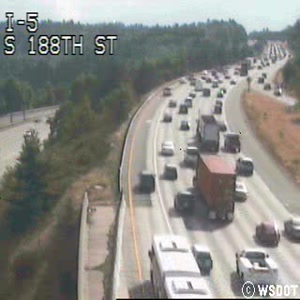}}
\subfloat[Heavy\label{H}]{%
        \includegraphics[width=0.33\linewidth]{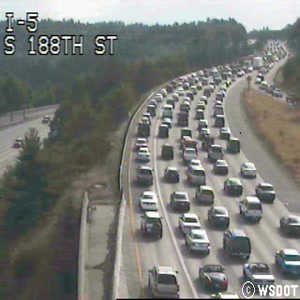}}
        \\
\subfloat[Rainy\label{1a}]{%
       \includegraphics[width=0.33\linewidth]{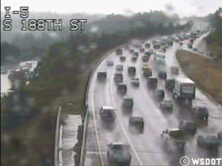}}
\subfloat[Foggy\label{1b}]{%
        \includegraphics[width=0.33\linewidth]{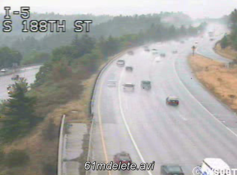}}
  \subfloat[Corrupted\label{1c}]{%
        \includegraphics[width=0.33\linewidth]{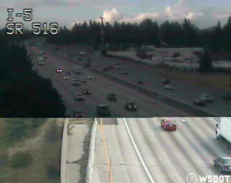}}
\caption{(Top row) dataset contains 3 modes of traffic jam; Low, Medium and Heavy. (Bottom row) shows poor quality and visibility due to precipitation or corrupted signal in the number of video samples.}
\label{classes} 
\end{figure}

The traffic video dataset contains 254 video samples of highway traffic in Seattle, recorded from a single
fixed traffic camera \cite{svcl}. The collection is divided into three classes, namely as Low, Medium and High traffic road congestion (Fig~\ref{classes}). Since the number of Low traffic jam label is considerably greater than the two others, the ``unbalanced'' samples happened in the dataset. The video samples also suffer from ``poor'' quality in the resolution of the recorded frames. Corrupted frames and sudden ``jumps'' between the frames are observed across the video frames. Moreover, a certain number of videos have been recorded in the ``precipitation'' weather condition where fog and rain affected on the visibility and clarity of the videos (Fig~\ref{classes}). These are the issues which have negative effects on the prediction. A certain number of videos have been recorded in the foggy or rainy climate, so the  visibility is drastically low in those videos. This is what causes the classification task more challenging, since the model should ignore noises due to water droplets spots and steam condition during the prediction process. 

\section{Model Selection}
\subsection*{YOLO}

\begin{figure*}[!ht]
\centering
\includegraphics[width=\textwidth]{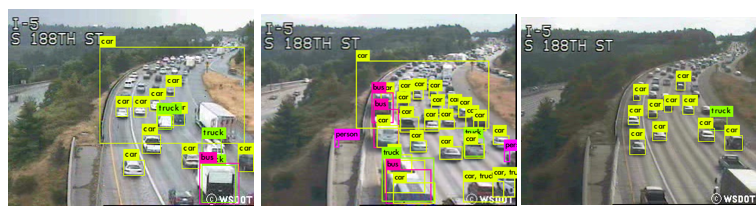}
\caption{Object detection in Videos using YOLO network. The middle scene clearly shows that YOLO recognizes Person and Truck by mistake, and it cannot detect vehicles in farther distance.}
\label{fig:yolodet}
\end{figure*}

YOLO (You Look Only Once) \cite{redmon2016you} developed an object detection framework as a regression problem to spatially separated bounding boxes and associated class probabilities. YOLO9000, a real-time object detection network which is the improved version of YOLO detection model, proposed by the same scientists, has fewer number of challenges that we had primarily by Google API. We deployed YOLO architecture using pre-trained weights to recognize objects across a video sample. Then we count the number of objects per frame (specifically vehicles) to classify each video into Low, Medium and Heavy on aggregated frames level, while the traffic flow is consistent across the frames. Several objects in the videos are quite clear, especially when there were “low” traffic congestion. So YOLO is able to label other objects in videos including trees, lanes and persons. The object ``Person'' is a false class YOLO specifies, since there is no person in the videos (Fig~\ref{fig:yolodet}). However, object detection task in Medium and Heavy classes are challenging, so YOLO fails to recognize more distinct objects as it does when traffic flow is Low (Fig~\ref{fig:yolodet}). This drawback is also seen in other Network architecture such as Google Video Intelligence Service which we address in the next section. 

% \begin{figure*}[!ht]
% \centering
% \includegraphics[width=\textwidth]{yolo.png}
% \caption{YOLO (You Only Look Once), a real-time object detection network architecture.}
% \label{fig:yolo}
% \end{figure*}

% ====================
% # obj detection
% ====================

% ====================
% # excel
% ====================
\begin{figure*}[!ht]
\centering
\includegraphics[width=\textwidth]{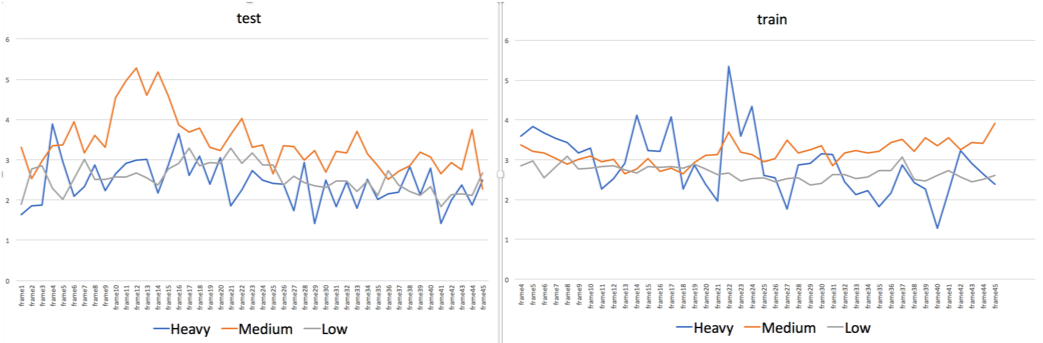}
\caption{Different variability in the number of detected vehicles in the test and train subsets.}
\label{fig:excel}
\end{figure*}

\subsection*{Google Video Intelligence API}
Another framework we analyzed performance is Google Cloud Video Intelligence API. There are cloud-based service providers such as Amazon, Google, Microsoft, BigML, and others who have developed MLAAS (Machine Learning As A Service) platform, so that individual users and commercial companies are able to employ pre-trained models to solve their problems without managing any hosts or servers. Also they can benefit from Machine Learning Engine models running on powerful machines contain CPU and GPU, so the problem solving will be accelerated tremendously. GUI in Google Video API is user-friendly, so this can be considered as an advantage. It is worth noting that Google API recognizes a traffic congestion directly from the videos, however, it does not specify the intensity of traffic. We observed that on the videos with Heavy or Medium traffic flow, the performance of YOLO architecture over Google API was particularly pronounced. So Google detects vehicles less than YOLO within the same video samples. In comparison with YOLO, Google API performs poorly in object detection task, and it fails when video samples are in foggy or rainy condition.

% \begin{figure}[!ht]
% \includegraphics[width=\linewidth]{google.png}
% \caption{Google Video Intelligence API architecture. Google has a nice GUI which enables user to run Machine Learning models easily.}
% \label{fig:google}
% \end{figure}

% ====================
% # CNN
% ====================
\section{Fresh CNN}
We build CNN models that can identify whether a given traffic flow video is predicted as Low, Medium or Heavy. Ultimately, we develop and train these CNN models from scratch to see how they would perform on new unseen data samples. These models are created in Keras framework, as a sequential model (Fig~\ref{fig:cnn}). The first layer would be a ``Convolutional Layer'' (Conv2D), and this is a 2-dimensional convolutional layer. The number of output filters in the convolution is 32, with $3\times3$ ``Kernel Size''. We use a ``Relu'' activation function, with $224\times224\times3$ specified in the first layer of sequential model for height, width and channel dimensions of frames, respectively. Each convolutional layer is followed by a ``Max Pooling'' layer with $2\times2$ pool size. We create CNNs with 5, 6 and 7 convolutional layers, and we have a flattened layer taking output from the previous layer and flattening it into a one dimensional tensor fed into a dense layer that has 3 nodes. The last layer contains 3 nodes since this will be the output layer that categorizing videos as Low, Medium or Heavy. We use the activation function of ``softmax'' in this last layer. The models are trained with ``Adam'' optimizer, learning rate = 0.00005, categorical cross-entropy loss function and an array with the single string accuracy as metrics. To prevent lengthy article, the ``optimization technique'' used in designing the CNNs architectures and choosing the semi-optimum hyper-parameters will be discussed in the future work. The summary of CNN models architectures are provided in Fig~\ref{fig:cnn}.

% \begin{figure}[!ht]
% \includegraphics[width=\linewidth]{cnn.png}
% \caption{CNN model architecture with 5 convolutional layers for traffic flow prediction.}
% \label{fig:cnn}
% \end{figure}

% ====================
% # VGG
% ====================
\section{Transfer Learning}
Using a pre-trained model for prediction is a growing technique practiced by many researchers. That model would achieve great results if the input data is classified into the categories used by the original model. However, none of the current pre-trained CNN architectures do not classify Highway Traffic Videos into specific "Mode" categories. Thus, one approach by which the scientists could obtain reasonable results is to Transfer the knowledge from one domain aspect to another. This is what people call as \textit{Transfer Learning}. Recent studies suggest that early layers of deep learning models identify simple ``patterns'', while later layers would identify more complex patterns. \cite{geirhos2018imagenet} showed that the ImageNet-trained CNNs are heavily biased in favor of recognising textures rather than shapes. Thus, the later layers in CNNs are complex representations of image textures. 
\subsection*{VGG19}
Very deep convolutional networks, also known as VGG, for large-scale image classification task is employed for labels prediction in traffic flow \cite{simonyan2014very}. The model increases depth using an architecture with
very small ($3\times3$) convolution filters which encourages higher performance in the localization and classification tasks. 

% \begin{figure*}[!ht]
% \centering
% \includegraphics[width=\textwidth]{vgg19.png}
% \caption{VGG19 model network architecture schema.}
% \label{fig:vgg}
% \end{figure*}

\subsection*{Model Setup}
Researchers in \cite{simonyan2014very} explained how well their CNN classifier model performed on classifying ImageNet challenge dataset. This section shows how VGG19 will be applied on a completely new type of dataset (CCTV videos of traffic flow) which does not contain classes similar to those included in ImageNet. VGG originally has been trained on images and now it is trained on videos. Video samples contain 3 classes labeled as Low (L), Medium (M) and Heavy (H) traffic flow, and each class is constructed of images of frames extracted from 5-second videos. This dataset is available on \cite{svcl} website as videos. We extracted the data as videos and extracted 10 frames per second. Having extracted and organized all frames from video samples, we created the directory iterators for train, validation and test sets. We built the new model that contains all of VGG19 layers up to its 5 to last layer, with an added output layer containing 3 output nodes that correspond to each of traffic mode classes. We observed that removing 2 fully connected layers (fc1 and fc2) would increase the accuracy of prediction significantly, while decreasing the number of ``Trainable Parameters'' and ``Total Parameters'' by 96\%. This DNN has been developed with Keras functional API framework. One question of interest is how many layers should be trained on the new dataset. One may still want to keep the mostly of what the original VGG19 model has already learned from ImageNet data by freezing the weights in the majority of layers. One solution is to implement ``Brute-Force'' approach on different number of layers, so one may find the optimal number as to unfreeze for training. We evaluated the performance of VGG19 on traffic videos by testing all the combinations of last 5 layers, and in our experiments, VGG tends to yield reasonably well accuracy with training the last 5 layers, which is ``block5\_conv1 (Conv2D)'' layer with output shape of (None, 14, 14, 512) and 2,359,808 parameters. All model configurations and results are provided in Table~\ref{Tab:vggcnn}. Under multiple settings, it achieved state-of-the-art results across several highly competitive configurations with the highest accuracy of 96.5\%.

% ====================
% # VGG table
% ====================
\begin{table*}
\centering
\captionof{table}{VGG19 and CNNs Table of Outputs\label{Tab:vggcnn}} 
\begin{tabular}{lcccc}
  \hline
 \textbf{Model} & \textbf{Hyper Parameters} & \textbf{Train} & \textbf{Validation} & \textbf{Test(CV)} \\ 
  \hline
 VGG & \{'trained\_layers': last 5, 'batch\_size': 100, 'steps\_per\_epoch: 45', 'epochs': '70', 'pre\_process': MobileNet\} & 100\% & 85.81\% & \textbf{96.50\%} \\ 
 VGG & \{'trained\_layers': last 5, 'batch\_size': 100, 'steps\_per\_epoch: 16', 'epochs': '70', 'pre\_process': MobileNet\} & 100\% & 85.91\% & \textbf{96.50\%} \\ 
 VGG & \{'trained\_layers': last 5, 'batch\_size': 287, 'epochs': '70', 'pre\_process': MobileNet\} & 100\% & 87.42\% & 95.42\% \\
 VGG & \{'trained\_layers': last 5, 'batch\_size': 300, 'epochs': '70', 'pre\_process': MobileNet\} & 100\% & 87.53\% & 95.27\% \\\hline
  CNN & \{'Conv2D\_layers': 5 layers, 'activation': Relu, 'epochs': '70', 'num\_dense\_nodes': 409\} & 100\% & 91.5\% & 92.84\% \\
   CNN & \{'Conv2D\_layers': 6 layers, 'activation': tanh, 'epochs': '70', 'num\_dense\_nodes': 449\} & 88.38\% & 83.08\% & 82.67\% \\
   CNN & \{'Conv2D\_layers': 7 layers, 'activation': Relu, 'epochs': '70', 'num\_dense\_nodes': 87\} & 100\% & 91.67\% & 91.98\% \\
   \hline
 & & \multicolumn{2}{c}{\textbf{state-of-the-art}} & 94.50\% \\
   \hline\\
\end{tabular}
\caption*{The results from VGG-Transfer-Learning approach under multiple configurations. The test results are calculated by the average of 10-fold cross validation. Several batch sizes included in the experiments to achieve higher performance. Conventionally, ``steps per epoch'' is set to an integer number obtained through division of train size over batch size. For batch size = 100, we set steps per epoch to 45 (simple division) and 16 (arbitrary number), to compare both settings. The table shows the highest accuracy is obtained by setting batch size to 100 and steps per epoch to 16.}
\end{table*}

% ====================
% # Results
% ====================
\section{Results}
Among 4 study cases, YOLO and Google Video Intelligence API are based on object detection technique. In other words, these techniques detect the number of objects (vehicles) per frame across a video and attempt to classify a given video based on average number of vehicles they recognize. As Fig~\ref{fig:yolodet} shows, both classifiers are not powerful enough to detect all the vehicles in a video, so classification merely based on the number of vehicles in the samples is not an appropriate idea. One reason could be video samples have not been recorded in high quality. Another reason is both classifiers are not able to segment between several vehicles when they are moving together, and they detect them mistakenly as truck or bus. Fig~\ref{fig:excel} shows that in Train plot, the average number of vehicles between Medium and Heavy modes are very close to one another, and the line in Heavy mode falls below the Medium line which is not correct. Interestingly, Heavy line (Blue) is below the Medium line (Orange) almost entirely. This technique may be useful when one would attempt to predict class labels between ``Low'' and ``non-Low'' modes, as Low lines (Grey) in Fig~\ref{fig:excel} indicate lower waving as opposed to other two lines (non-Low). In our experiments, YOLO tends to yield better performance in detecting vehicles than Google Video Intelligence API by segmenting objects more reliably, with less number of false negative. Also we observed that YOLO is more robust against sever conditions than Google Video Intelligence API, where the weather is foggy or rainy in videos. However, Google Video Intelligence API has a richer GUI and user can benefit from uploading the video on Google Storage to run models faster than a local system, simply because Google runs the models in a big data framework and a high scalable infrastructure. The CNN and VGG-Transfer-Learning appear to be an effective way to classify videos' contents. The results presented in Table~\ref{Tab:vggcnn} reveal that VGG19 outperforms all the fresh CNNs in prediction task. In the same table, the CNN with 5 convolutional layers gained the highest accuracy (92.84\%). Although CNN-5 achieves a relatively high accuracy in classification (92\%), VGG19 obtains 96.5\% accuracy which is 4.5\% more than CNN-5. CNN-5 misclassifies 77 frames videos as Medium mode, while they are Heavy, without any misclassification between Low and Medium or Low and Heavy. In VGG19, on the other hand, several hyper-parameters have been tested and evaluated. We observed that when the batch size decreases from 300 to 100, the number of misclassifications will also drop significantly (50\%). The misclassification happens when predicting between Medium and Heavy modes. Under multiple settings, it achieved state-of-the-art results with training last 5 layers and 70 epochs (Table~\ref{Tab:vggcnn}).

% ====================
% # Performance?
% ====================
\section{Model Performance}
Tables~\ref{tab:vggsens}~\&~\ref{tab:cnnsens} are vividly showing that both CNN-5 and VGG19 classifiers performed perfectly well in classifying Low Vs. non-Low classes, as it shows values 100\% for Sensitivity and Specificity. However, both models are less efficient in predicting between Medium and Heavy classes. Although Sensitivity has been recorded very high for CNN (100\%) and VGG19 (97.54\%), Specificity needs improvement especially in the CNN-5 model. In other words, the table reports that CNN-5 would be able to identify 83\% of Heavy Traffic flow video cases as Heavy correctly, and classify 17\% as Medium Traffic flow incorrectly. Similarly, VGG can classify almost 91\% of Heavy Traffic flow cases as Heavy correctly, and predict the rest (9\%) as Medium incorrectly.
\newline 
The results presented in Table~\ref{tab:vggsens}~\&~\ref{tab:cnnsens} reveal that VGG-Transfer-Learning achieved the largest Accuracy, Sensitivity and Specificity results across several highly competitive hyper-parameters with nearly perfect evaluation metrics, while CNN-5 performed on par with VGG in Sensitivity criteria.

% ====================
% # VGG-sens-spec
% ====================
\begin{table}[!htbp]
\renewcommand{\arraystretch}{1.2}
\setlength{\arrayrulewidth}{0.1mm}
\setlength{\doublerulesep}{0.1mm}
\caption{VGG-Transfer-Learning Diagnostic Table.}
\label{tab:vggsens}
\centering
\begin{tabular}{|c|c|c|c|c|}
\hline
\multicolumn{1}{|c|}{\textbf{Pair-Labels}}  &\textbf{\#batches}  &  \textbf{Sensitivity} &  \textbf{Specificity} & \textbf{Accuracy}  \\
\hline
\multirow{3}{*}{Low-Medium} & 100 & 100\% & 100\% & 100\% \\
& 287 & 100\% & 100\% & 100\%  \\
& 300 & 100\% & 100\% & 100\%\\ \hline
\multirow{3}{*}{Low-Heavy} & 100 & 100\% & 100\% & 100\%\\
& 287 & 100\% & 100\% & 100\%\\
& 300 & 100\% & 100\% & 100\%\\ \hline
\multirow{3}{*}{Medium-Heavy} & 100 & 97.12\% & 94.42\% & 95.74\%\\
& 287 & 97.54\% & 91.12\% & 94.12\%\\
& 300 & 97.54\% & 90.74\% & 93.91\%\\ \hline
\end{tabular}
\end{table}

% ====================
% # CNN-sens-spec
% ====================
\begin{table}[!htbp]
\renewcommand{\arraystretch}{1.2}
\setlength{\arrayrulewidth}{0.1mm}
\setlength{\doublerulesep}{0.1mm}
\caption{Diagnostic Table for the Best CNN-5 (5 Conv2D).}
\label{tab:cnnsens}
\centering
\begin{tabular}{|c|c|c|c|}
\hline
\multicolumn{1}{|c|}{\textbf{Pair-Labels}} &  \textbf{Sensitivity} &  \textbf{Specificity} & \textbf{Accuracy}  \\
\hline
\multirow{1}{*}{Low-Medium} & 100\% & 100\% & 100\%\\ \hline
\multirow{1}{*}{Low-Heavy} & 100\% & 100\% & 100\%\\ \hline
\multirow{1}{*}{Medium-Heavy} & 100\% & 83.71\% & 89.71\%\\ \hline
\end{tabular}
\end{table}

% ====================
% # Why VGG?
% ====================
\section{Why VGG19 outperforms?}
\subsection*{Transfer Values}
To extract the transfer values, we exclude the last layer in VGG architecture, which is the ``softmax'' classification layer and we call it ``Transfer Layer'' and its output is ``Transfer Values''. We store this layer to hard disk, since this is an expensive computation. The transfer values are nothing but arrays with 25,088 elements due to the output shape of VGG-Transfer-Learning architecture. Thus, the transfer values have $1396\times25088$ dimensions, with 1396 and 25088 represent the number of samples (video frames) and features, respectively. This new space contains pertinent information as to how they form classes in a high dimension space.

% ====================
% # PCA
% ====================

\subsection*{Dimension Reduction: PCA}
\begin{figure}[!ht]
\includegraphics[width=\linewidth]{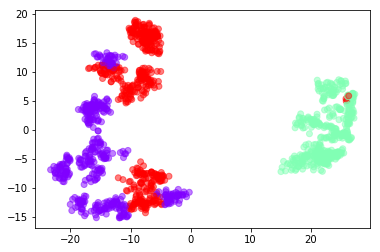}
\caption{The dimension reduction technique using PCA method to reduce the dimension size from 25088 to 2. Three colors refer to three class labels (L, M, H). This plot illustrates information flow throughout VGG network up to ``bottleneck'' layer.}
\label{fig:pca}
\end{figure}

\begin{figure}[!ht]
\includegraphics[width=\linewidth]{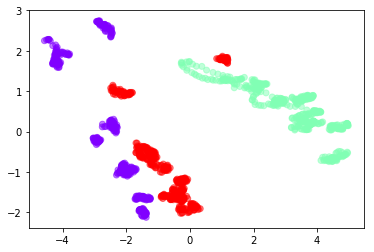}
\caption{The dimension reduction technique using PCA method to reduce the dimension size from 12544 to 2. Three colors refer to three class labels (L, M, H). This plot illustrates information flow throughout CNN with 5 convolutional layers network up to ``bottleneck'' layer.}
\label{fig:cnn5pca}
\end{figure}

As part of this study, we are interested in analyzing the transfer values, so we can learn how the VGG19 model is able to extract useful information and separate 3-class labels we specify. The challenge is the transfer values are obtained in a very high dimension space (25088 and 12544 elements in VGG19 and CNN-5, respectively), so plotting is impossible.
For ease of exposition, we will apply a widely-used dimension reduction technique, called PCA. We call this method from ``Scikit-learn'' package with n = 2 meaning all transfer values are reduced to arrays of length 2. The PCA process is implemented for the entire test size (1396 samples), each of them is an array with 25088 and 12544 values, the dimension size of VGG19 and CNN-5 networks, respectively. Ultimately, the PCA is reduced from 25088 and 12544 values to only 2 values. Fig~\ref{fig:pca}~\&~\ref{fig:cnn5pca} show PC1 and PC2 for the transfer values in VGG19 (top) and CNN-5 (bottom). The plots illustrate visibly that class Low (green) is easily separable from other two classes, while Medium and Heavy classes are mixing at some points. The class colors Red and Purple might be Medium and Heavy classes, or vice versa. As the plots suggest, the VGG19-Transfer-Learning model that we train on traffic flow video samples would predict as expect, since the information extracted from video frames are properly separated into 3 classes with minor overlapping between Red and Purple. However, the transfer values in CNN-5 tend to be distributed more sparsely after dimension reduction, which results in difficulty for the classifier to perform classification. Fig~\ref{fig:pca}~\&~\ref{fig:cnn5pca} are in line with the results on classification performance we provided in Table~\ref{tab:vggsens}~\&~\ref{tab:cnnsens}. We benefit from PCA technique since it is linear and deterministic, but other dimension reduction technique such as ``t-SNE'' (t-Distributed Stochastic Neighbor Embedding) which is non-linear and non-deterministic is also applicable, however, t-SNE would not perform well when the data samples are relatively large.
% ==================
% # PCAs #
% ==================

% ==================
% # Conclusion #
% ==================
\section{Conclusion}
We presented several experiments in understanding contents of videos. Two approaches, YOLO9000 and Google Video Intelligence API, were depending on object detection fashion, and other two, CNN and VGG-Transfer-Learning, predicted classes through convolutional paradigm. We showed that video contents understanding scales naturally to tens of frames and objects inside, while exhibiting no implementation difficulties. Whilst object detection approach follows a simple object counting rule in the frame-level,
distinguishing between two similar classes (Medium and Heavy) are challenging. In our experiments, YOLO outperformed Google Video Intelligence API in detecting more vehicles across video frames, while Google was more user friendly. We also trained and evaluated a new CNN from scratch to compare it with another pre-trained model (VGG19). Although the CNN model performed astonishingly well, VGG showed higher accuracy in prediction task due to its complex network architecture and large number of examples fed as input. It is conceivable that more extensive hyper-parameter searches may further improve the performance of CNN on Traffic database. Under multiple settings, VGG19 achieved the state-of-the-art. In our experiments, both CNN architectures tended to yield consistent results in accuracy with reducing number of parameters, without any signs of performance degradation. On the traffic video dataset, we observed that the simplicity of our approaches over prior work are particularly pronounced. The results presented in Table~\ref{Tab:vggcnn} revealed that our best solution (VGG19) performed better than the state-of-the-art KL-SVM (98.50\% $>$ 94.5\%), whilst requiring significantly fewer parameters tuning and computation to achieve supremacy.

% ==================
% # Acknowledgment #
%\section*{Acknowledgment}
%For the Summary paper submission only, no %acknowledgements are allowed. 

% ==============
% # REFERENCES #
% ==============
\bibliographystyle{IEEEtran}
\bibliography{IEEEabrv,biblio_rectifier}

% ===============================================
%\section*{\textbf{Appendix}}
% ==================
% #CNN plots
% ==================
\begin{figure*}[!ht]
\minipage{0.34\textwidth}
  \includegraphics[width=\textwidth]{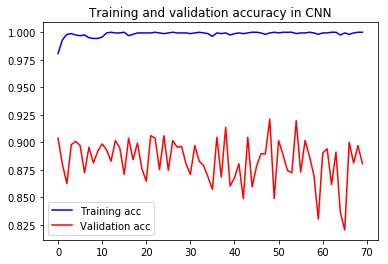}
  \caption*{Training \& Validation Accuracy in CNN with 5 convolutional layers and batch\_size = 100.}\label{fig:cnnAcc}
\endminipage\hfill
\minipage{0.34\textwidth}
  \includegraphics[width=\textwidth]{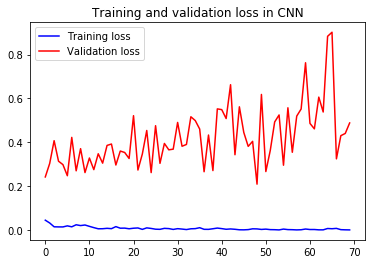}
  \caption*{Training \& Validation Loss in CNN with 5 convolutional layers and batch\_size = 100.}\label{fig:cnnLoss}
\endminipage\hfill
\minipage{0.27\textwidth}%
  \includegraphics[width=\textwidth]{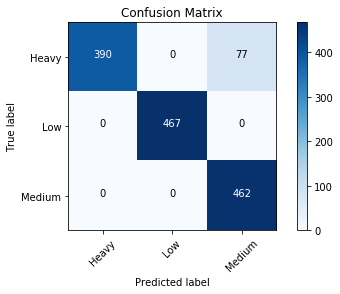}
  \caption*{Confusion Matrix in CNN with 5 convolutional layers and batch\_size = 100. }\label{fig:cmCNN100}
\endminipage
\end{figure*}

% ==================
% #VGG100-16 plots
% ==================
\begin{figure*}[!ht]
\minipage{0.34\textwidth}
  \includegraphics[width=\textwidth]{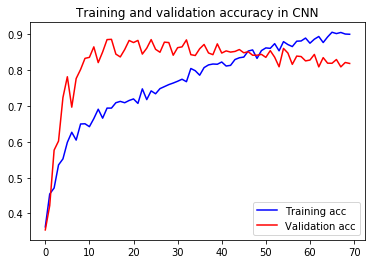}
  \caption*{Training \& Validation Accuracy in VGG with batch\_size = 100 and steps\_per\_epoch = 16.}\label{fig:cnnAcc}
\endminipage\hfill
\minipage{0.34\textwidth}
  \includegraphics[width=\textwidth]{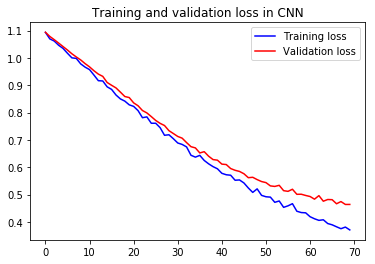}
  \caption*{Training \& Validation Loss in VGG with batch\_size = 100 and steps\_per\_epoch = 16.}\label{fig:cnnLoss}
\endminipage\hfill
\minipage{0.27\textwidth}%
  \includegraphics[width=\textwidth]{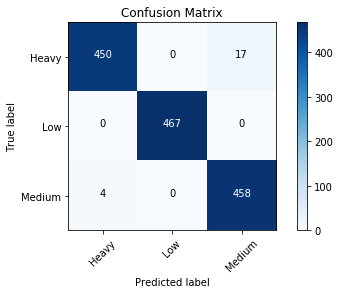}
  \caption*{Confusion Matrix in VGG with batch\_size = 100 and steps\_per\_epoch = 16.}\label{fig:cm100}
\endminipage
\end{figure*}

% ==================
% #VGG287 plots
% ==================
\begin{figure*}[!ht]
\minipage{0.34\textwidth}
  \includegraphics[width=\textwidth]{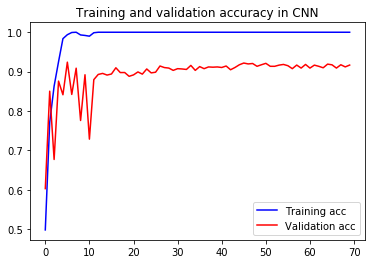}
  \caption*{Training \& Validation Accuracy in VGG with batch\_size = 287.}\label{fig:cnnAcc}
\endminipage\hfill
\minipage{0.34\textwidth}
  \includegraphics[width=\textwidth]{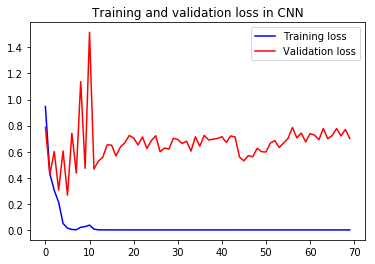}
  \caption*{Training \& Validation Loss in VGG with batch\_size = 287.}\label{fig:TV287loss}
\endminipage\hfill
\minipage{0.27\textwidth}%
  \includegraphics[width=\textwidth]{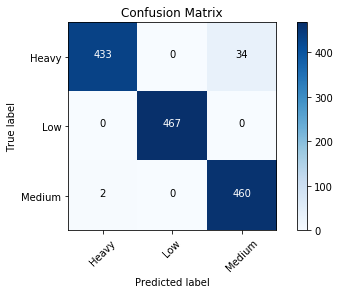}
  \caption*{Confusion Matrix in VGG batch\_size = 287.}\label{fig:cm287}
\endminipage
\end{figure*}

\end{document}